\documentclass[11pt]{article}
\usepackage{acl2016}
\usepackage{times}
\usepackage{url}
\usepackage{latexsym}
\usepackage{amsmath}
\usepackage{amssymb}
\usepackage{array}
\usepackage{algorithm}
\usepackage[noend]{algpseudocode}
\usepackage{booktabs}
\usepackage{tikz}
\usepackage{forest}
\usepackage{subcaption}

\makeatletter
\def\BState{\State\hskip-\ALG@thistlm}
\makeatother

\makeatletter
\newcommand{\algmargin}{\the\ALG@thistlm}
\makeatother
\newlength{\whilewidth}
\algdef{SE}[parWHILE]{parWhile}{EndparWhile}[1]
  {\parbox[t]{\dimexpr\linewidth-\algmargin}{%
     \hangindent\whilewidth\strut\algorithmicwhile\ #1\ \algorithmicdo\strut}}{\algorithmicend\ \algorithmicwhile}%
\algnewcommand{\parState}[1]{\State%
  \parbox[t]{\dimexpr\linewidth-\algmargin}{\strut #1\strut}}

\aclfinalcopy % Uncomment this line for the final submission

\title{The AMU-UEDIN Submission to the WMT16 News Translation Task: \\ Attention-based NMT Models as Feature Functions in Phrase-based SMT}

\author{Marcin Junczys-Dowmunt\textsuperscript{1,2}, Tomasz Dwojak\textsuperscript{1}, Rico Sennrich\textsuperscript{2} \\
\textsuperscript{1}Faculty of Mathematics and Computer Science, Adam Mickiewicz University in Pozna\'{n}\\
\textsuperscript{2}School of Informatics, University of Edinburgh \\
{\tt \normalsize junczys@amu.edu.pl t.dwojak@amu.edu.pl}\\
{\tt \normalsize rico.sennrich@ed.ac.uk }}

\date{}

\begin{document}
\maketitle
\begin{abstract}
This paper describes the AMU-UEDIN submissions to the WMT 2016 shared task on news translation. We explore methods of decode-time integration of attention-based neural translation models with phrase-based statistical machine translation. Efficient batch-algorithms for GPU-querying are proposed and implemented. 
For English-Russian, our system stays behind the state-of-the-art pure neural models in terms of BLEU. Among restricted systems, manual evaluation places it in the first cluster tied with the pure neural model. 
For the Russian-English task, our submission achieves the top BLEU result, outperforming the best pure neural system by 1.1 BLEU points and our own phrase-based baseline by 1.6 BLEU. After manual evaluation, this system is the best restricted system in its own cluster. In follow-up experiments we improve results by additional 0.8 BLEU.
\end{abstract}

\section{Introduction}

This paper describes the AMU-UEDIN submissions to the WMT 2016 shared task on news translation. We explore methods of decode-time integration of attention-based neural translation models with phrase-based decoding. Experiments have been conducted for the English-Russian language pair in both translation directions. 

For these experiments we re-implemented the inference step of the models described in  \newcite{DBLP:journals/corr/BahdanauCB14} (more exactly the DL4MT\footnote{\url{https://github.com/nyu-dl/dl4mt-tutorial}} variant also present in Nematus\footnote{\url{https://github.com/rsennrich/nematus}}) in efficient C++/CUDA code that can be directly compiled as a Moses feature function. The GPU-based computations come with their own peculiarities which we reconcile with the two most popular phrase-based decoding algorithms --- stack-decoding and cube-pruning.

While it seems at first that for English-Russian our phrase-based system is holding back the neural models in terms of BLEU, the manual evaluation reveals that our systems is tied with the pure neural systems, occupying the same top cluster for restricted systems with an even slightly higher TrueSkill score.
We achieve the top BLEU result for the Russian-English task, outperforming the best pure neural system by 1.1 BLEU points and our own phrase-based baseline by 1.6 BLEU. After manual evaluation, this system is the best restricted system in its own cluster.

Our implementation is available as a Moses fork from \url{https://github.com/emjotde/mosesdecoder_nmt} 

\section{Preprocessing}

As we reuse the neural systems from \newcite{sennrich-wmt16}, we follow their preprocessing scheme for the phrase-based systems as well. All data is tokenized with the Moses tokenizer, for English the Penn-format tokenization scheme has been used. Tokenized text is true-cased.

\newcite{sennrich-wmt16} use byte-pair-encoding (BPE) to achieve open-vocabulary translation with a fixed vocabulary of subword symbols \cite{DBLP:journals/corr/SennrichHB15}. For English, the vocabulary size is limited to 50,000 units, for Russian to 100,000. This has the interesting consequence of using subword units for phrase-based SMT. Although SMT seems to be better equipped to handle large vocabularies, the case of Russian still poses problems which are usually solved with transliteration mechanisms \cite{DBLP:conf/eacl/DurraniSHK14}. Resorting to subword units eliminates the need for these.\footnote{In experiments not described in this paper, we tried BPE encoding for the English-German language pair and found subword units to cope well with German compound nouns when used for phrase-based SMT.}

\section{Neural translation systems}

As mentioned before, we reuse the English-Russian and Russian-English NMT models from \newcite{sennrich-wmt16} and refer the reader to that paper for a more detailed description of these systems. In this section we give a short summarization for the sake of completeness. 

The neural machine translation system is an attentional encoder-decoder \cite{DBLP:journals/corr/BahdanauCB14}, which has been trained with Nematus.
Additional parallel training data has been produced by automatically translating a random sample (2 million sentences) of the monolingual Russian News Crawl 2015 corpus into English \cite{2015arXiv151106709S}, which has been combined with the original parallel data in a 1-to-1 ratio.\footnote{This artificial data has not been used for the creation of the phrase-based system, but it might be worthwhile to explore this possibility in the future. It might enable the phrase-based system to produce translation that are more similar to the neural output.} The same has been done for the other direction.
We used mini-batches of size 80, a maximum sentence length of 50, word embeddings of size 500, and hidden layers of size 1024.
We clip the gradient norm to 1.0 \cite{DBLP:conf/icml/PascanuMB13}.
Models were trained with Adadelta \cite{DBLP:journals/corr/abs-1212-5701}, reshuffling the training corpus between epochs.
The models have been trained model for approximately 2 weeks, saving every $30000$ mini-batches. 

For our experiments with PB-SMT integration, we chose the same four models that constituted the best-scoring ensemble from \newcite{sennrich-wmt16}. If less than four models were used, we chose the models with the highest BLEU scores among these four models as measured on a development set. 

\section{Phrase-Based baseline systems}

We base our set-up on a Moses system \cite{Koehn:2007:MOS:1557769.1557821} with a number of additional feature functions.
Apart from the default configuration with a lexical reordering model, we add a 5-gram operation sequence model \cite{conf/acl/DurraniFSHK13}. 

We perform no language-specific adaptations or modifications. The two systems differ only with respect to translation direction and the available (monolingual) training data.
For domain-adaptation, we rely solely on parameter tuning with Batch-Mira \cite{Cherry:2012:BTS:2382029.2382089} and on-line log-linear interpolation. Binary domain-indicators for each separate parallel corpus are introduced to the phrase-tables (four indicators) and a separate language model per parallel and monolingual resource is trained (en:16 and ru:12). All language models are 5-gram models with Modified Kneser-Ney smoothing and without pruning thresholds \cite{Heafield-estimate}. We treat different years of the News Crawl data as different domains to take advantage of possible recency-based effects. During parameter tuning on the newstest-2014 test set, we can unsurprisingly observe that weights for the last three LMs (2013, 2014, 2015) are much higher than for the remaining years. 

After concatenating all resources, a large 5-gram background language model is trained, with 3-grams or higher n-gram orders being pruned if they occur only once. The same concatenated files and pruning settings are used to create a 9-gram word-class language model with 200 word-classes produced by word2vec \cite{journals/corr/abs-1301-3781}. 

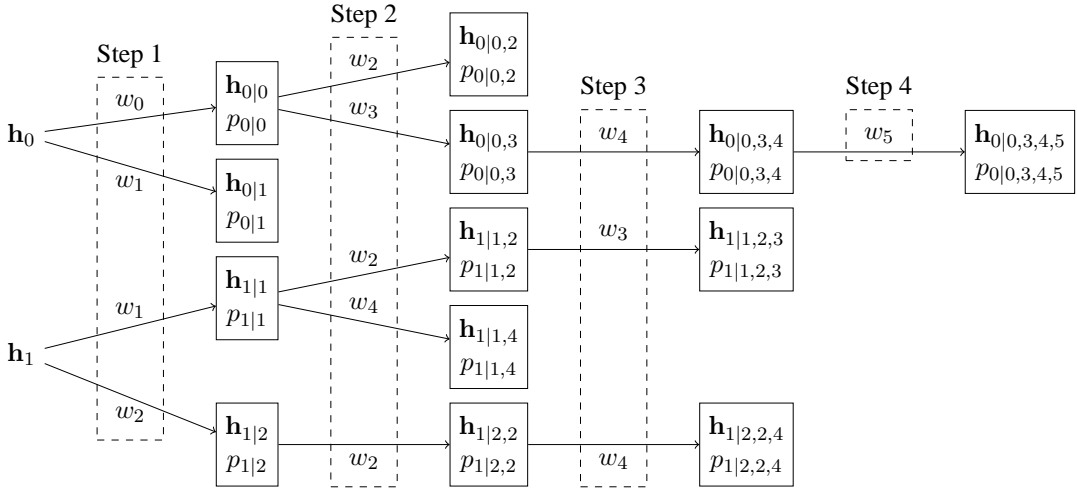
\begin{figure*}[t!]
\begin{center}
\scalebox{0.9}{
\begin{forest}
  phantom,
  for tree={
    grow'=east, s sep=0.2cm, l sep=2.5cm,
    edge={->},
  }
  [[$\mathbf{h}_0$
    [$\mathbf{h}_{0|0}$\\$p_{0|0}$, rectangle,draw,align=center, edge label={node[midway,above](s1_1) {$w_0$}} 
      [$\mathbf{h}_{0|0,2}$\\$p_{0|0,2}$, rectangle,draw,align=center,edge label={node[midway,above](s2_1){$w_2$}}]
      [$\mathbf{h}_{0|0,3}$\\$p_{0|0,3}$, rectangle,draw,align=center,edge label={node[midway,above](s2_2){$w_3$}}
	[$\mathbf{h}_{0|0,3,4}$\\$p_{0|0,3,4}$, rectangle,draw,align=center,edge label={node[midway,above](s3_1){$w_4$}}
	  [$\mathbf{h}_{0|0,3,4,5}$\\$p_{0|0,3,4,5}$, rectangle,draw,align=center,edge label={node[midway,above](s4_1){$w_5$}}]
	]
      ]
    ]
    [$\mathbf{h}_{0|1}$\\$p_{0|1}$, rectangle,draw,align=center,edge label={node[midway,below](s1_2){$w_1$}}]
  ]
  [$\mathbf{h}_1$
    [$\mathbf{h}_{1|1}$\\$p_{1|1}$, rectangle,draw,align=center,edge label={node[midway,above](s1_3){$w_1$}} 
      [$\mathbf{h}_{1|1,2}$\\$p_{1|1,2}$, rectangle,draw,align=center,edge label={node[midway,above](s2_3){$w_2$}}
	[$\mathbf{h}_{1|1,2,3}$\\$p_{1|1,2,3}$, rectangle,draw,align=center,edge label={node[midway,above](s3_2){$w_3$}}]]
      [$\mathbf{h}_{1|1,4}$\\$p_{1|1,4}$, rectangle,draw,align=center,edge label={node[midway,above](s2_4){$w_4$}}]
    ]
    [$\mathbf{h}_{1|2}$\\$p_{1|2}$, rectangle,draw,align=center, edge label={node[midway,below](s1_4){$w_2$}}
      [$\mathbf{h}_{1|2,2}$\\$p_{1|2,2}$, rectangle,draw,align=center,edge label={node[midway,below](s2_5){$w_2$}}
	[$\mathbf{h}_{1|2,2,4}$\\$p_{1|2,2,4}$, rectangle,draw,align=center,edge label={node[midway,below](s3_3){$w_4$}}]
      ]
    ]
  ]]
  \node[draw,dashed,fit=(s1_1) (s1_2) (s1_3) (s1_4),label=Step 1] {} ;
  \node[draw,dashed,fit=(s2_1) (s2_2) (s2_3) (s2_4) (s2_5),label=Step 2] {};
  \node[draw,dashed,fit=(s3_1) (s3_2) (s3_3),label=Step 3] {};
  \node[draw,dashed,fit=(s4_1),label=Step 4] {};
\end{forest}}
\end{center}
\caption{\textsc{ScoreBatch} procedure for a forest consisting of two per-hypothesis prefix trees. Words are collected at the same tree depths across all trees in the forest.}\label{alg:viz}
\end{figure*}

\section{NMT as Moses feature functions}

As mentioned in the introduction, we implemented a C++/CUDA version of the inference step for the neural models trained with DL4MT or Nematus, which can be used directly with our code. One or multiple models can be added to the Moses log-linear model as different instances of the same feature, which during tuning can be separately weighted. Adding multiple models as separate features becomes thus similar to ensemble translation with pure neural models. 

In this section we give algorithmic details about integrating GPU-based soft-attention neural translation models into Moses as part of the feature function framework.
Our work differs from \newcite{alkhouli-rietig-ney:2015:WMT} in the following aspects:
\begin{enumerate}
 \item While \newcite{alkhouli-rietig-ney:2015:WMT} integrate RNN-based translation models in phrase-based decoding, this work is to our knowledge the first to integrate soft-attention models. 
 \item Our implementation is GPU-based and our algorithms being tailored towards GPU computations require very different caching strategies from those proposed in \newcite{alkhouli-rietig-ney:2015:WMT}. Our implementation seems to be about 10 times faster on one GPU, 30 times faster when three GPUs are used. 
\end{enumerate}

\subsection{Scoring hypotheses and their expansions}

We assume through-out this section that the neural model has already been initialized with the source sentence and that the source sentence context is available at all time. 

In phrase-based machine translation, a pair consisting of a translation hypothesis and a chosen possible target phrase that expands this hypothesis to form a new hypothesis can be seen as the smallest unit of computation. In the typical case they are processed independently from other hypothesis-expansion pairs until they are put on a stack and potentially recombined. Our aim is to run the computations on one or more GPUs. This makes the calculation of scores per hypothesis-expansion pair (as done for instance during n-gram language model querying) unfeasible as repeated GPU-access with very small units of computation comes with a very high overhead. 

In neural machine translation, we treat neural states to be equivalent to hypotheses, but they are extended only by single words, not phrases, by performing computations over the whole target vocabulary. In this section, we present a batching and querying scheme that aims at taking advantage of the capabilities of GPUs to perform batched calculations efficiently, by combining the approaches from phrase-based and neural decoding. 

Given is a set of pairs $(h,t)$ where $h$ is a decoding hypothesis and $t$ a target phrase expanding the hypothesis. In a naive approach (corresponding to unmodified stack decoding) the number of queries to the GPU would be equal to the total number of words in all expansions. A better algorithm might take advantage of common target phrase prefixes per hypothesis. The number of queries would be reduced to the number of collapsed edges in the per-hypothesis prefix-tree forest. 

By explicitly constructing this forest of prefix trees where a single prefix tree encodes all target phrases that expand the same hypothesis, we can actually reduce the number of queries to the neural model to the maximum depth of any of the trees (i.e. the maximum target phrase length) as illustrated in Figures \ref{alg:viz} and \ref{alg1}.

\begin{figure}[t]
\begin{algorithmic}[1]
\Procedure{ScoreBatch}{$L$, \textsc{NMT}}
\parState {Create forest of per-hypothesis prefix trees from all hypotheses and expansions in $L$}
\For{$i$ \textbf{from} $1$ \textbf{to} $\text{maximum tree depth}$}
\parState{Construct embedding matrix $E_i$ from all edge labels at depth $i$} 
\parState{Construct row-wise corresponding state matrix $H_{i-1}$ from source nodes}
\parState{Compute forward step:\\ $(H_i, P_i) \gets \textsc{Nmt}(H_{i-1}, E_i)$}
\parState{Cache state pointers and probabilities at target nodes}
\EndFor
\EndProcedure
\end{algorithmic}
\caption{Scoring of hypothesis expansion pairs}
\label{alg1}
\end{figure}

Target phrases $t$ are treated as sequences of words $w$. Rectangles at tree nodes should be imagined to be empty before the preceding step has been performed. The first embedding matrix $E_1$ is constructed by concatenating embedding vectors $\textbf{e}_i \gets \textsc{Lookup}(w_i)$ as rows of the matrix, for all $w_i$ marked in the first dashed rectangle. The initial state matrix $H_0$ is a row-wise concatenation of the neural hypothesis states, repeated for each outgoing edge. Thus, the embedding matrix and state matrix have the same number of corresponding rows. Example matrices for the first step take the following form:

$$
E_1 = \left[\begin{array}{c}\mathbf{e}_0 \\ \mathbf{e}_1  \\ \mathbf{e}_1 \\ \mathbf{e}_2\end{array}\right] \quad
H_0 = \left[\begin{array}{c}\mathbf{h}_0 \\ \mathbf{h}_0  \\ \mathbf{h}_1 \\ \mathbf{h}_1\end{array}\right] \quad
$$

Given the precomputed source context state, we can now perform one forward step in the neural network which yields a matrix of output states and a matrix of probabilities, both corresponding row-wise to the input state matrix and embedding matrix we constructed earlier. The target nodes for each edge pointed to after the first step are filled. Probabilities will be queried later during phrase-based scoring, neural hypothesis states are reused to construct the state matrix of the next step and potentially as initial states when scoring another batch of hypotheses at later time.

\subsection{Two-pass stack decoding}

\begin{figure}[t]
\begin{algorithmic}[1]
\Procedure{TwoPassStackDecoding}{}
\State Place empty hypothesis $h_0$ into stack $S_0$
\For{stack $S$ \textbf{in} stacks }
   \State $L \gets \emptyset$
   \State \Call{ProcessStack}{$S$, \textsc{Gather\{$L$\}}}
   \State $C \gets $ \Call{ScoreBatch}{$L$, \textsc{Nmt}}
   \State \Call{ProcessStack}{$S$, \textsc{Expand\{$C$\}}}
\EndFor
\EndProcedure\\
\Procedure{ProcessStack}{$S$, $f$}
  \For{hypothesis $h$ \textbf{in} $S$}
    \For{target phrase $t$}
      \If{applicable}
	\State Apply functor \Call{$f$}{$h, t$}
      \EndIf
    \EndFor
  \EndFor
\EndProcedure\\
\Procedure{Gather}{$h$, $t$}
  \State $L \gets L \cup \{(h,t)\}$
\EndProcedure\\

\Procedure{Expand}{$h$, $t$}
\State Look-up $p$ for $(h,t)$ in $C$
\State Create new hypothesis $\hat{h}$ from $(h,t,p)$
\State Place $\hat{h}$ on appropriate stack $s$
\If{possible} \parState{Recombine hypothesis $\hat{h}$ with other hypotheses on stack $s$}
\EndIf
\If{stack $s$ too big} \State Prune stack $s$
\EndIf
\EndProcedure
\end{algorithmic}
\caption{Two-pass stack decoding}\label{alg2}
\end{figure}

Standard stack decoding still scores hypotheses one-by-one. In order to limit the number of modifications of Moses to a minimum, we propose two-pass stack decoding where the first pass is a hypothesis and expansions collection step and the second pass is the original expansion and scoring step. Between the two steps we pre-calculate per-hypothesis scores with the procedure described above. The data structure introduced in Figure~\ref{alg:viz} is then reused for probability look-up during the scoring phrase of stack decoding as if individual hypotheses where scored on-the-fly.

Figure~\ref{alg2} contains our complete proposal for two-pass stack decoding, a modification of the original stack decoding algorithm described in \newcite{Koehn:2010:SMT:1734086}. We dissect stack decoding into smaller reusable pieces that can be passed functors to perform different tasks for the same sets of hypotheses. The main reason for this is the small word ``applicable'' in line 12, which hides a complicated set of target phrase choices based on reordering limits and coverage vectors which should not be discussed here. This allows our algorithm to collect exactly the set of hypotheses and expansions for score pre-calculation that will be used during the second expansion step. 

As already mentioned, the number of forward steps for the NMT network per stack is equal to the greatest phrase length among all expansions. The total number of GPU queries increases therefore linearly with respect to the sentence length. Branching factors or stack sizes affect the matrix sizes, not the number of steps.\footnote{Large matrix sizes, however, do slow-down translation speed significantly.}

For this method we do not provide results due to a lack of time. We confirmed for other experiments that improvements are smaller than for the next method. A comparison will be provided in an extended version of this work.

\subsection{Stack rescoring}

The previous approach cannot be used with lazy decoding algorithms --- like cube pruning --- which has also been implemented in Moses. Apart from that, due to the large number of expansions even small stack sizes of around 30 or 50 quickly result in large matrices in the middle steps of \textsc{BatchScore} where the prefix trees have the greatest number of edges at the same depth level. In the worst case, matrix size will increase by a factor $b^d$, where $b$ is the branching factor and $d$ is the current depth. In practice, however, the maximum is reached at the third or fourth step, as only few target phrases contain five or more words. 

\begin{figure}[t]
\begin{algorithmic}[1]
\Procedure{StackRescoring}{}
\State Place empty hypothesis $h_0$ into stack $S_0$
\For{stack $S$ \textbf{in} stacks }
   \State $L \gets \emptyset$
   \For{hypothesis $h$ \textbf{in} $S$}
      \State Extract predecessors $(\bar{h}, \bar{t})$ from $h$
      \State $L \gets L \cup \{(\bar{h}, \bar{t})\}$
   \EndFor
   \State $C \gets $ \Call{ScoreBatch}{$L$, \textsc{Nmt}}
   \For{hypothesis $h$ \textbf{in} $S$}
      \State Extract predecessors $(\bar{h}, \bar{t})$ from $h$
      \State Look-up $p$ for $(\bar{h}, \bar{t})$ in $C$
      \State Recalculate score of $h$ using $p$
   \EndFor
   \State Create cache $C_0$ with 0-probabilities
   \State \Call{ProcessStack}{$S$, \textsc{Expand}\{$C_0$\}}
\EndFor
\EndProcedure
\end{algorithmic}
\caption{Stack decoding with stack rescoring}\label{alg3}
\end{figure}

To address both shortcomings we propose a second algorithm: stack rescoring. This algorithm (Figure~\ref{alg3}) relies on two ideas:
\begin{enumerate}
 \item During hypothesis expansion the NMT feature is being ignored, only probabilities of 0 are assigned for this feature to all newly created hypotheses. Hypothesis recombination and pruning take place without NMT scores for the current expansions (NMT scores for all previous expansions are included). Any stack-based decoding algorithm, also cube-pruning, can be used in this step. 
\item The \textsc{BatchScore} procedure is applied to all direct predecessors of hypotheses on the currently expanded stack. Predecessors consist of the parent hypothesis and the expansion that resulted in the current hypothesis. The previously assigned 0-probabilities are replaced with the actual NMT scores. 
\end{enumerate}

\begin{table*}[t]
 \centering 
 \subcaptionbox{BLEU scores English-Russian}[0.48\textwidth]{
\centering 
\begin{tabular}{lcc}\toprule
System & 2015  & 2016 \\ \midrule
Phrase-Based (PB) & 23.7 & 22.8 \\ \midrule
Pure neural:\\
%NMT-1 (Sennrich et al. 2016)& 25.8 & 24.3 \\
NMT-2 & 26.4 & 25.3 \\
NMT-4 \cite{sennrich-wmt16} & 27.0 & 26.0 \\ 
%\midrule
%1000-best list re-ranking: & tbc & tbc\\
%\midrule
%Two-pass stack decoding: & tbc & tbc\\
\midrule
Stack rescoring:\\ 
%PB+NMT-1 & 25.5 & tbc \\
\textbf{PB+NMT-2 (subm.)} & --- & \bf 25.3 \\  \midrule \midrule
Follow-up:\\ 
NMT-4-Avg & 26.7 & 25.5 \\
PB+NMT-4-Avg & 27.3 & 25.9 \\ \bottomrule
\end{tabular}
\label{tab:enru}}
\subcaptionbox{BLEU scores Russian-English}[0.48\textwidth]{
\centering 
\begin{tabular}{lcc}\toprule
System &  2015  & 2016 \\ \midrule
Phrase-Based (PB) &   27.4 & 27.5 \\ \midrule
Pure neural: \\
%NMT-1 (Sennrich et al. 2016) & 27.1 & 26.9  \\
NMT-3 & 28.3 & 27.8\\
NMT-4 \cite{sennrich-wmt16} & 28.3 & 28.0 \\
%\midrule
%1000-best list re-ranking: & tbc & tbc\\
%\midrule
%Two-pass stack decoding: & tbc & tbc \\
\midrule
Stack rescoring:\\ 
%PB+NMT-1 & tbc & tbc \\
\textbf{PB+NMT-3 (subm.)} & \bf 29.5 & \bf 29.1 \\ \midrule \midrule
Follow-up:\\ 
NMT-10-Avg & 28.3 & 28.1 \\ 
\bf PB+NMT-10-Avg & \bf 30.2 & \bf 29.9 \\ \bottomrule
\end{tabular}
\label{tab:ruen}}
\caption{Systems marked with \textbf{subm.} are our final WMT 2016 submissions. %Some contrastive results are still being computed, we mark these with ``tbc'' and plan to supply them for the camera-ready version.
} \label{tab:compare}
\end{table*}

This procedure results in a number of changes when compared to standard stack decoding approaches and the previous method:

\begin{itemize}
 \item The maximum matrix row count is equal to the stack size, and often much smaller due to prefix collapsing. Branching factors are irrelevant and stack sizes of 2,000 or greater are possible. By contrast, for two-pass stack decoding stack sizes of around 50 could already result in row counts of 7,000 and more.
\item With cube pruning, by setting cube pruning pop-limits much larger than the stack size many more hypotheses can be scored with all remaining feature functions before the survivors are passed to \textsc{BatchScore}.
\item Scoring with the NMT-feature is delayed until the next stack is processed. This may result in missing good translations due to recombination. However, the much larger stack sizes may counter this effect. 
\item N-best list extraction is more difficult, as hypotheses that have been recombined do not display correct cumulative sums for the NMT-feature scores. The one-best translation is always correctly scored as it has never been discarded during recombination, so there is no problem at test time. For tuning, where a correctly scored n-best list is required, we simply rescore the final n-best list with the same neural feature functions as during decoding. The resulting scores are the same as if they were produced at decode-time. 
Final n-best list rescoring can thus be seen as an integral part of stack-rescoring. 
\end{itemize}

\section{Experiments and results}

For decoding, we use the cube-pruning algorithm with stack size of 1,000 and cube-pruning pop limit of 2,000 during tuning. At test time, a stack-size of 1,000 is kept, but the cube-pruning pop limit is increased to 5,000. We set a distortion limit of 12. We run 10 iterations of Batch-Mira \cite{Cherry:2012:BTS:2382029.2382089} and choose the best set of weights based on the development set. Our development set is a subset of 2,000 sentences from the newstest-2014 test set. Sentences have been selected to be shorter than 40 words to avoid GPU-memory problems. 
Our GPUs are three Nvidia GeForce GTX-970 cards with 4GB RAM each. 

In this paper, similar as \newcite{alkhouli-rietig-ney:2015:WMT}, we ignore the implications of the infinite neural state and hypothesis recombination in the face of infinite state. We rely on the hypothesis recombination controlled by the states of the other feature functions. It is worth mentioning again that our phrase-based baseline features a 9-gram word-class language model which should be rather prohibitive of recombinations. If recombination was only allowed for hypotheses with the same partial translations, results were considerably worse. 

\subsection{Speed}

\begin{table}[t]
\centering
 \begin{tabular}{lr}
 \toprule
 & words/s \\ \midrule
\newcite{alkhouli-rietig-ney:2015:WMT} (1 thread?) &  0.19 \\ 
%Phrase-based (1 thread) &  \\
Phrase-based PB (24 threads) & 40.30 \\ 
%PB-NMT-1, Stack rescoring, 1 GPU & 1.61 \\
PB-NMT-10-Avg (3 GPUs) & 4.83\\
\bottomrule
 \end{tabular}
 \caption{Translation speed for different configurations in words per second.}
 \label{tab:speed}
\end{table}

Translation speed is difficult to compare across systems (Table~\ref{tab:speed}). Even with three GPUs our system is ten times slower than than a pure PB-SMT system running with 24 CPU-threads. It is however unclear at this moment if the large stack sizes we use are really necessary. Significant speed-up might be achieved for smaller stacks.

\subsection{Submitted results}

Table~\ref{tab:compare} summarizes the results for our experiments. BLEU scores are reported for the newstest-2015 and newstest-2016 test sets.

Our baseline phrase-based systems (PB) are quite competitive when comparing to the best results of last year's WMT (24.4 and 27.9 for English-Russian and Russian-English, respectively). NMT-4 is the best pure neural ensemble from \newcite{sennrich-wmt16} for both translation directions. Due to memory restrictions, we were not able to use all four models as separate feature functions and limit ourselves to the best two models for English-Russian and best three for Russian-English. The pure neural ensembles are NMT-2 (en-ru) and NMT-3 (ru-en), respectively. 

For English-Russian, our results stay behind the pure-neural 4-ensemble NMT-4 in terms of BLEU. In a direct comparison between ensembles of 2 models (PB+NMT-2 and NMT-2), we actually reach similar BLEU scores. However, in the manual evaluation our system is best restricted system, tied with the neural system. Absolute TrueSkill scores are even slightly higher for our system. 

For Russian-English the best-performing pure neural system NMT-4 and the phrase-based baseline are only 0.5\% BLEU apart. Adding three NMT models as feature functions to Moses results in a 1.1\% BLEU improvement over the neural model and 1.6\% over the phrase-based system. The systems PB-NMT-2 (en-ru) and PB-NMT-3 (ru-en) are our submissions to the WMT-2016 news translation task. PB-NMT-3 scores the top BLEU results for Russian-English. In the manual evaluation, our system is the best restricted system in its own cluster.

\subsection{Follow-up experiments}

Frustrated by the limited memory of our GPU cards and against better knowledge\footnote{The neural network lore seems to suggest that this should not work, as neural networks are non-linear models. We only found one paper with evidence to the contrary: \newcite{Utans96weightaveraging}}, we computed the  element-wise average of all model weights in the NMT ensembles and saved the resulting model. Interestingly, the performance of these new models (NMT-4-Avg) is not much worse than the actual ensemble (NMT-4), while being four times smaller and four times faster at decode-time. The average models outperforms any single model or the smaller 2-ensembles. All models taking part in the average are parameter dumps saved at different points in time during the same training run. This seem to be an interesting results for model compression and deployment settings. We can also average more models: for the Russian-English direction we experiment with the parameter-wise average of ten models (NMT-10-Avg) which even slightly outperforms the real four-model ensemble NMT-4. 

With this smaller model it is easier to tune and deploy our feature function. The performance of our combined setup improves for both translation directions. For English-Russian, however, the pure NMT system (NMT-4) remains ahead of our WMT 2016 submission. For Russian-English we get another improvement of 0.8 BLEU, which sets the new state-of-the-art for this direction.

\section*{Acknowledgments}
 
This project has received funding from the European Union's Horizon 2020 research and innovation
programme under grant agreements 644333 (TraMOOC) and 688139 (SUMMA) and was partially funded by the Amazon Academic Research Awards programme.

\bibliography{acl2016}
\bibliographystyle{acl2016}

\end{document}